\documentclass[letterpaper]{article} % DO NOT CHANGE THIS
\usepackage{aaai2026}  % DO NOT CHANGE THIS
\usepackage{times}  % DO NOT CHANGE THIS
\usepackage{helvet}  % DO NOT CHANGE THIS
\usepackage{courier}  % DO NOT CHANGE THIS
\usepackage[hyphens]{url}  % DO NOT CHANGE THIS
\usepackage{graphicx} % DO NOT CHANGE THIS
\usepackage{natbib}  % DO NOT CHANGE THIS AND DO NOT ADD ANY OPTIONS TO IT
\usepackage{caption} % DO NOT CHANGE THIS AND DO NOT ADD ANY OPTIONS TO IT
\usepackage{algorithm}
\usepackage{algorithmic}

\usepackage{newfloat}
\usepackage{listings}
\DeclareCaptionStyle{ruled}{labelfont=normalfont,labelsep=colon,strut=off} % DO NOT CHANGE THIS
\floatstyle{ruled}
\newfloat{listing}{tb}{lst}{}
\floatname{listing}{Listing}
\usepackage{amsmath}
\usepackage{amsfonts}
\usepackage{amssymb}
\usepackage{booktabs}
\usepackage{multirow}

\title{EPO: Diverse and Realistic Protein Ensemble Generation via Energy Preference Optimization}
\author{
    Yuancheng Sun\textsuperscript{\rm 1, \rm 2, \rm 3},
    Yuxuan Ren\textsuperscript{\rm 3},
    Zhaoming Chen\textsuperscript{\rm 3},
    Xu Han\textsuperscript{\rm 1, \rm 2, \rm 3},
    Kang Liu\textsuperscript{\rm 1, \rm 2},
    Qiwei Ye\textsuperscript{\rm 3}\thanks{Corresponding Author.}
}
\affiliations{
    \textsuperscript{\rm 1}Institute of Automation, Chinese Academy of Sciences\\
    \textsuperscript{\rm 2}University of Chinese Academy of Sciences\\
    \textsuperscript{\rm 3}Beijing Academy of Artificial Intelligence\\
    sunyuancheng2021@ia.ac.cn,
    chivee.ye@gmail.com
}

\begin{document}

\maketitle

\begin{abstract}
Accurate exploration of protein conformational ensembles is essential for uncovering function but remains hard because molecular-dynamics (MD) simulations suffer from high computational costs and energy-barrier trapping.
This paper presents Energy Preference Optimization (EPO), an online refinement algorithm that turns a pretrained protein ensemble generator into an energy-aware sampler without extra MD trajectories.
Specifically, EPO leverages stochastic differential equation sampling to explore the conformational landscape and incorporates a novel energy-ranking mechanism based on list-wise preference optimization.
Crucially, EPO introduces a practical upper bound to efficiently approximate the intractable probability of long sampling trajectories in continuous-time generative models, making it easily adaptable to existing pretrained generators.
On Tetrapeptides, ATLAS, and Fast-Folding benchmarks, EPO successfully generates diverse and physically realistic ensembles, establishing a new state-of-the-art in nine evaluation metrics.
These results demonstrate that energy-only preference signals can efficiently steer generative models toward thermodynamically consistent conformational ensembles, providing an alternative to long MD simulations and widening the applicability of learned potentials in structural biology and drug discovery.
\end{abstract}

% \begin{links}
%     \link{Code}{https://github.com/sunyuancheng/EPO}
% \end{links}

\section{Introduction}
Proteins operate through continual interconversion among multiple conformational states.
This dynamic landscape underlies fundamental biological processes, including allostery, molecular recognition, and catalysis~\cite{nussinov2016introduction, raisinghani2024probing, henzler2007dynamic}.
Consequently, mechanistic insight~\cite{nussinov2023protein,kalakoti2025afsample2, schafer2025sequence}—and, by extension, structure-based drug discovery—requires models that recover the entire Boltzmann ensemble rather than a single static conformation~\cite{ teixeira2022idpconformergenerator, nussinov2016introduction, raisinghani2024probing}.

\begin{figure}[t]
    \centering
    \includegraphics[width=\linewidth]{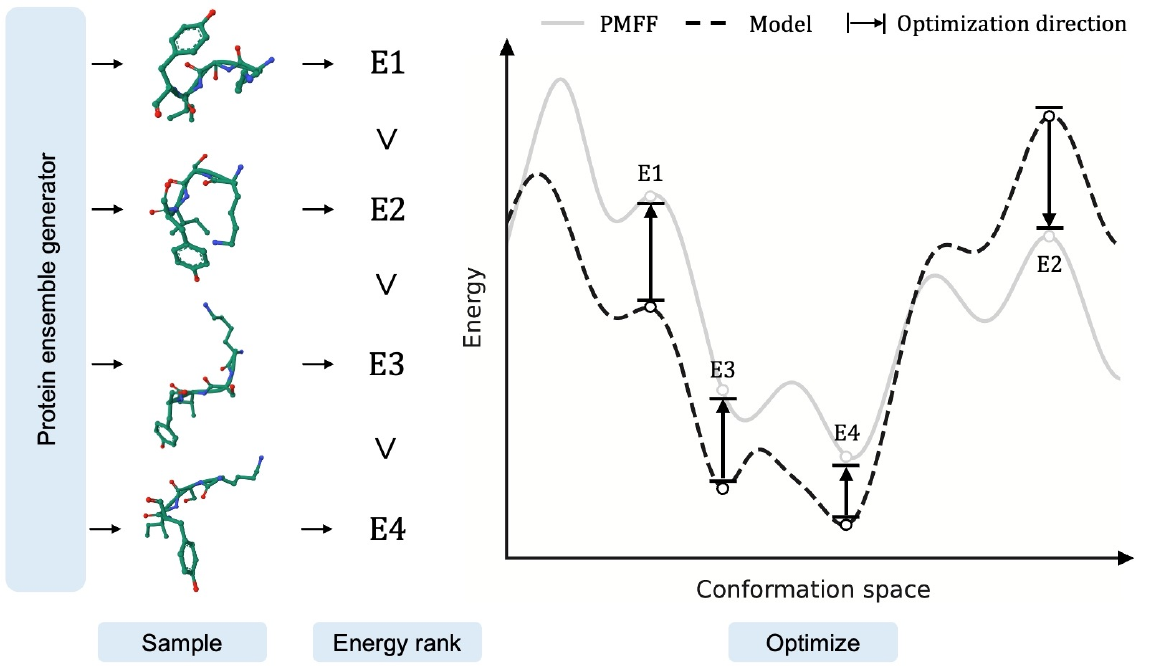}
    \caption{EPO aligns the model's energy landscape with the ground truth by leveraging energy rankings from a physics-based molecular force field (PMFF), guiding the model to assign lower energy to more favorable conformations.}
    \label{fig:epo-intro}
    % \vspace{-4mm}
\end{figure}

Molecular dynamics (MD) remains the standard tool for exploring these ensembles~\cite{stevens2023molecular,alrawashdeh2023applications,ghahremanian2022molecular,badar2022molecular}. 
Yet MD trajectories are constrained by rugged energy landscapes whose high barriers render functional transitions rare on simulation timescales~\cite{henin2022enhanced, ray2022rare, wang2021gaussian,souza2021martini}.
Conventional runs therefore become trapped in local minima and undersample transient or high-energy states that are often functionally critical.

Recent advancements in deep generative models offer a computationally efficient approach to achieving conformational diversity.
Early efforts perturb single-state predictors such as AlphaFold2~\cite{jumper2021highly} with dropout, template shuffling, or MSA subsampling~\cite{del2022sampling, stein2022speach_af, wayment2024predicting}, but yield only modest variations near the dominant conformation. 
A more prevalent strategy involves fine-tuning generative models on extensive MD trajectories~\cite{bioemu, oc2, klein2024transferable}. 
However, supervised fine-tuning may suffer from misalignment out of inadequate sampling. 
This issue is two-fold.
First, producing equilibrated MD trajectories necessitates prohibitively long simulations that may still undersample the true ensemble in large biomolecular systems~\cite{henin2022enhanced, ray2022rare}.
Second, even when fine-tuned on well-sampled trajectories, generative models tend to revisit the same local regions of conformational space, leading to generated conformations that fail to accurately follow the MD-distribution~\cite{confdiff, mdgen}.

Recent advances in large language models have revealed the effectiveness of post-training refinement techniques in addressing the misalignment issues in supervised fine-tuning~\cite{wallace2024diffusion,guo2025deepseek}.
Direct Preference Optimization (DPO)~\cite{dpo} stands out as a representative example of such methods.
Drawing a parallel, we posit that pretrained protein ensemble generators can be viewed as competent yet misaligned samplers. 
However, directly applying DPO to these generators is non-trivial. 
Ensemble generation aims to reproduce the full Boltzmann distribution of relevant states, whereas the pairwise comparisons used in naïve DPO tend to steer the model toward a single, energetically favorable basin, undermining distributional fidelity~\cite{wang2023beyond,han2025f,lanchantin2025diversepreferenceoptimization}.

To address these challenges, this paper introduces Energy Preference Optimization (EPO), a novel physics-guided online framework for refining protein ensemble generators.
Specifically, EPO integrates stochastic differential equation (SDE) sampling into an online, iterative refinement process.
This allows the model to dynamically explore the conformational landscape, thereby mitigating the issue of inadequate sampling.
Furthermore, to overcome the limitations of pairwise comparisons, EPO introduces a physics-based energy ranking mechanism. 
This mechanism employs listwise preference optimization to guide the generator towards a diverse and physically realistic ensemble rather than a single low-energy state. 
Crucially, EPO derives a practical upper bound for the listwise preference objective, effectively approximating the intractable transition probabilities of long sampling trajectories inherent in continuous-time generative models.
This formulation integrates seamlessly with modern denoising score-matching and flow-matching architectures and can be adopted directly by existing pretrained generators.

Extensive experiments demonstrate that preference signals derived solely from physical energy are sufficient to correct the misalignment of pretrained generators. 
As a result, this relieves the need for expensive post-hoc MD trajectories.
Empirically, EPO establishes a new state-of-the-art across nine distinct distributional metrics on the Tetrapeptides~\cite{mdgen}, ATLAS~\cite{atlas}, and Fast-Folding~\cite{fastfolding} benchmarks.
Furthermore, visualizations and ablation studies validate the effectiveness of EPO's design components.

Our contributions are summarized as follows:
\begin{itemize}
    \item This paper proposes EPO, an online framework that directly aligns pretrained generators with the target Boltzmann distribution using listwise energy preferences.
    \item A practical upper bound for the intractable listwise preference objective is derived, yielding tractable gradients compatible with modern continuous-time generative models.
    % \item Extensive experiments show EPO achieves state-of-the-art performance on multiple benchmarks, producing diverse and physically realistic ensembles without requiring additional MD simulations.
    \item Extensive experiments show that EPO attains state-of-the-art results on several metrics and benchmarks, and delivers competitive performance elsewhere, while producing diverse and physically realistic ensembles without requiring additional MD simulations.
\end{itemize}

\begin{figure*}[t]
    \centering
    \includegraphics[width=\textwidth]{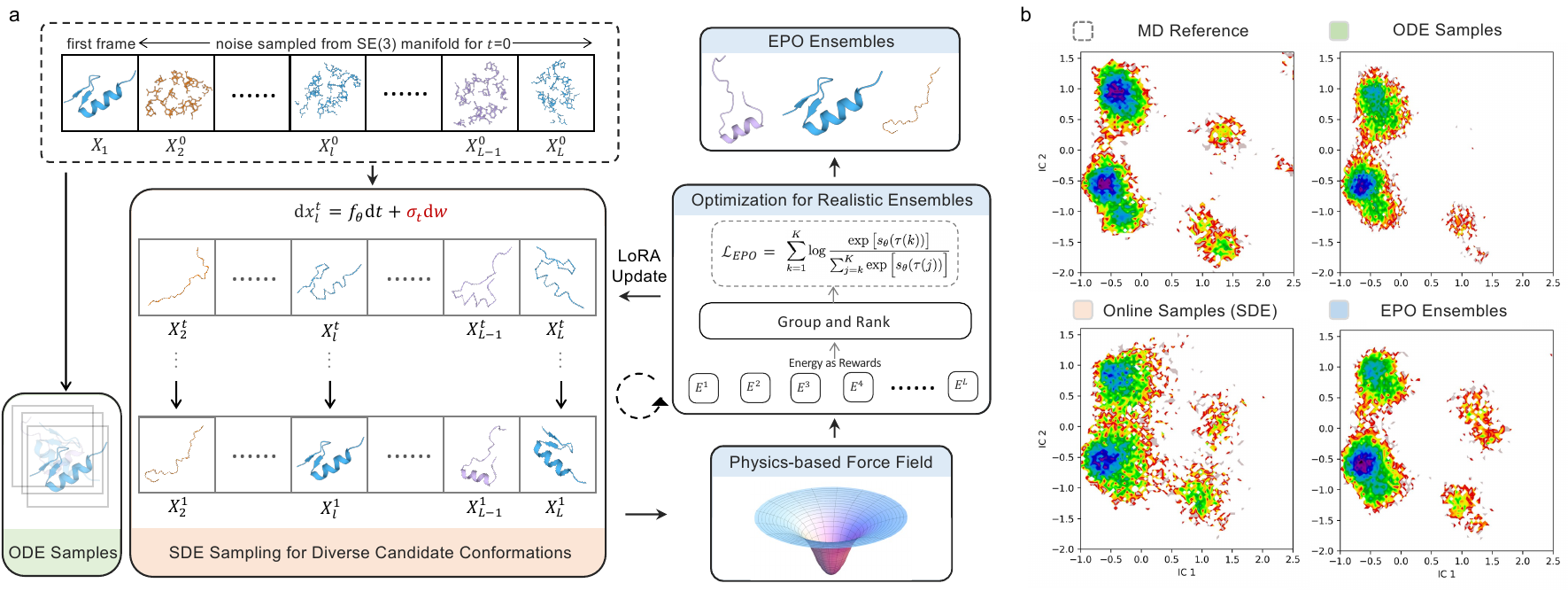}
    \caption{Overview of EPO. (a) Given the first frame (static conformation) of a protein, we leverage a pretrained ensemble generator and introduce an ODE-to-SDE strategy to enable stochastic sampling for online optimization. Energies of the online samples are used as rewards to update the model with LoRA. (b) SDE-based sampling effectively overcomes the energy barriers inherent in the original model (sequence: ASRE). Consequently, the optimized EPO model generates diverse and physically realistic protein ensembles, bypassing the need for any post-hoc MD simulations.}
    \label{fig:overview}
% \vspace{-5mm}
\end{figure*}
% \vspace{-3mm}

\section{Preliminaries}
\paragraph{Direct Preference Optimization.}
Let a prompt $x\!\in\!\mathcal{X}$ be given and let a policy
$\pi_\theta(y|x)$ assign probabilities to responses
$y \!\in\! \mathcal{Y}$ (e.g.\ molecular conformations, textual answers).
The learning goal is to adjust $\theta$ so that the induced distribution matches externally provided preferences.

Direct Preference Optimization (DPO)~\cite{dpo} learns from
a dataset
$\mathcal{D}= \{(x^{(i)},y^{(i)}_{w},y^{(i)}_{l})\}_{i=1}^{N}$,
where each triple contains a winning sample
$y_{w}$ and a losing sample $y_{l}$.
Under the Bradley–Terry comparison model, the DPO loss is
% \vspace{-2mm}
\begin{multline}
    \label{eq:dpo_bt}
    \mathcal{L}_{\text{DPO-BT}}(\pi_\theta; \pi_{\text{ref}}) =  - \mathbb{E}_{(x, y_w, y_l) \sim \mathcal{D}}\Big[ \\
     \log \sigma \big( \beta \log \frac{\pi_\theta(y_w|x)}{\pi_{\text{ref}}(y_w|x)}
     - \beta \log \frac{\pi_\theta(y_l|x)}{\pi_{\text{ref}}(y_l|x)} \big) \Big]
\end{multline}
where $\sigma$ is the logistic function and
$\beta\!>\!0$ controls the implicit KL regularization with respect to a fixed reference policy $\pi_{\text{ref}}$, i.e., the initial pretrained checkpoint.

In practice, human preferences often extend beyond pairwise comparisons to ranked lists, introducing a key challenge that the magnitude of preference between items is typically non-uniformly weighted. 
Consequently, when a prompt is associated with an ordered list
$\tau=(y_{(1)},\dots,y_{(m)})$ sorted by preference,
we can generalize Eq.~\ref{eq:dpo_bt} to a listwise variant,
which can be derived from the Plackett–Luce model~\cite{dpo,lipo} or from ListMLE~\cite{listmle}:
% \vspace{-2mm}
\begin{multline}
\label{eq:dpo_pl}
\mathcal{L}_{\text{DPO-PL}}(\pi_\theta; \pi_{\text{ref}}) = - \mathbb{E}_{(\tau,x,y_1,...,y_k) \sim \mathcal{D}} \Big[ \\
 -\sum_{i=1}^K \log \frac{\exp(s_\theta(x^{\tau(i)}))}{\sum_{j=i}^K \exp(s_\theta(x^{\tau(j)}))} \Big],
\end{multline}
where $s_\theta(x^i) \overset{\Delta}{=} \beta\log \frac{\pi_\theta(y|x^i)}{\pi_{\text{ref}}(y|x^i)}$.
% and setting $m=2$ recovers Eq.~\ref{eq:dpo_bt}.

% Its success in aligning large language models motivates our adoption of a listwise analogue for protein conformation generation.

\paragraph{Flow Matching and SDE Sampling.}
Flow Matching (FM)~\cite{flowmatching,albergo2022building} trains a time-dependent velocity network $v_\theta(x,t)$ that transports samples from a simple prior distribution $p_0(x)$ (e.g., $\mathcal{N}(0,I)$) to a target data distribution $p_1(x)$. 
This process is defined by constructing a coupling path between pairs of samples $(x_0, x_1) \sim p_0 \times p_1$. 
For any such pair, the path $x_t$ for $t \in [0,1]$ is defined as:
$x_t = \alpha_t x_1 + \sigma_t x_0$.
Here, $\alpha_t$ and $\sigma_t$ are scheduling functions designed to ensure the path smoothly interpolates from the prior sample to the data sample.
% For instance, a common choice is linear interpolation, where $\alpha_t = t$ and $\sigma_t = 1-t$, satisfying the boundary conditions $x_0 \sim p_0(x)$ at $t=0$ and $x_1 \sim p_1(x)$ at $t=1$.
Let $\dot{x}_t=\dot{\alpha}_t x_1 + \dot{\sigma}_t\epsilon$ denote the ground-truth velocity along this path.
FM fits $v_\theta$ by minimizing the expectation of a mean squared error (MSE) loss:
\begin{equation}
\label{eq:flowmatching}
  \mathcal{L}_{\mathrm{FM}}(\theta)\;
  =\;
  \mathbb{E}_{x_0,x_1,t}\!
  \bigl[
     \lVert v_\theta(x_t,t)-\dot{x}_t \rVert_2^{2}
  \bigr].
\end{equation}
After training, samples can be generated by solving a probability flow ordinary differential equation (ODE):
$x_1 = \texttt{odeint}(v_\theta(t),x_0,t\!:\!0\!\rightarrow\!1)$.
Following \citeauthor{ma2024sit}, we construct a reverse-time SDE formulation that preserves the same marginal distribution as the ODE:
% \vspace{-1mm}
\begin{equation}
dx_t = v(x_t, t)dt + \frac{1}{2}w_t s(x_t, t)dt + \sqrt{w_t}d\bar{\mathbf{W}}_t,    
\end{equation}
% \vspace{-1mm}
\begin{equation}
\label{eq:score-velocity-transform}
s(x_t,t)=\sigma_{t}^{-1} \frac{\alpha_tv_\theta(x_t,t)-\dot{\alpha}_tx_t}{\dot{\alpha}_t\sigma_t-\alpha_t\dot{\sigma_t}},
\end{equation}
where $d\bar{\mathbf{W}}_t$ denotes the Wiener process and $w_t>0$ is the score norm controlling the level of stachasticity.

This equivalence allows us to employ existing pretrained velocity networks within an SDE sampling framework to enhance sampling diversity and aid optimization, as has been demonstrated in recent literature~\cite{str2str,xue2025dancegrpounleashinggrpovisual, liu2025flowgrpotrainingflowmatching}.

\paragraph{FlowDPO.}
For autoregressive language or diffusion models, $\log\pi_\theta(y|x)$ is directly available as token logits or trajectory log-probabilities, making DPO an efficient, black-box-free alternative to RLHF.
Unlike language models, closed-form $\log\pi_\theta(y|x)$ of FM is computational ;therefore, the pairwise DPO loss (Eq.~\ref{eq:dpo_bt}) appears intractable because it requires
trajectory-wide log-likelihoods.
FlowDPO~\cite{flowdpo} resolves this by amortizing the pairwise DPO objective over time:
under a Gaussian assumption on marginal states, the global preference loss decomposes into a weighted difference of per-time \(\text{MSE}_t(x_0,x_1;\theta)\), 
each of which is computable from $v_\theta$ in Eq.~\ref{eq:flowmatching} alone:
% \vspace{-2mm}
\begin{multline}
\label{eq:dpo_fm}
    \mathcal{L}_{\text{DPO-FM}} = -\mathbb{E}_{x^w_{0,1},x^l_{0,1},t}\log \sigma ( \\
    \beta \big[ \text{MSE}_t(x^w_0, x^w_1; \theta_{\text{ref}}) - \text{MSE}_t(x^w_0, x^w_1; \theta_{\text{opt}}) \\
    - \text{MSE}_t(x^l_0, x^l_1; \theta_{\text{ref}}) + \text{MSE}_t(x^l_0, x^l_1; \theta_{\text{opt}}) ] \big).
\end{multline}
% \vspace{-1mm}
A detailed proof can be found in Appendix.

\section{Method}
Prevailing pairwise preference optimization techniques fall short in modeling complex systems, as they cannot capture the free energy hierarchy of the conformational ensemble. 
Listwise objectives offer a more powerful alternative but are typically computationally intractable. 
To bridge this gap, we introduce EPO, a new framework that leverages a practical upper bound on the listwise objective. 
This formulation reframes the problem as a direct energy minimization task, unlocking the potential of the listwise approach by ensuring both its effectiveness and computational feasibility.

\subsection{Suboptimality of Pairwise DPO for Energy Alignment}
Generally, a protein free-energy landscape is characterized by several metastable basins whose local minima may lie at markedly different absolute energies.
When a generator is trained with pairwise Direct Preference Optimization (Eq.~\ref{eq:dpo_bt}), every comparison favors the conformation of lower energy, gradually steering the model toward the single, globally best basin.
This pairwise pressure undermines the very goal of ensemble modelling—namely, to reproduce the full distribution of relevant states—because higher-energy but kinetically important basins are progressively ignored.
A listwise loss, which updates the parameters with respect to the ordering of an entire batch, treats these basins more even-handedly and therefore better preserves diversity.

Beyond diversity, listwise objectives possess a functional form—typically a log-sum-exp over batch scores (Eq.~\ref{eq:dpo_pl})—that can be interpreted as soft importance reweighting.
Low-energy samples receive exponentially larger gradients, yet the contribution of higher-energy samples does not vanish entirely.
Recent work on energy-based fine-tuning has shown that such adaptive weighting is beneficial for matching the target Boltzmann distribution~\cite{uehara2025inference,li2024derivative,confdiff}.

Finally, the computational route to parity between the two settings is impractical.
Although evaluating all $\binom{K}{2}$ comparisons in each batch allows a pairwise objective to be equivalent to a listwise objective, this is prohibitive and impractical for realistic values of $K$ in flow-matching models.
Restricting the loss to a subset of pairs—such as adjacent elements after energy sorting—reduces cost but at the price of higher gradient variance and information loss.
This simplified training setting is denoted as EPO-Pair and its underperformance compared to the full listwise version, EPO-List, is demonstrated in the experimental section.

For these reasons, EPO is designed with a listwise structure to guide the generative model towards the underlying physical energy landscape in a manner that is both robust and diversity-promoting.

\subsection{Extension on Flow Matching}
Adapting the Listwise Preference Optimization (LiPO) framework, originally designed for discrete outputs from language models, to continuous-time generative models like flow matching presents a significant challenge.
The core of this adaptation lies in redefining the reward mechanism. 
While standard LiPO evaluates a model's final, static output, the continuous dynamics of flow matching necessitate a reward defined over the entire generation trajectory.
Therefore, drawing inspiration from recent work on preference optimization for generative models~\cite{wallace2024diffusion,flowdpo}, we propose a trajectory-based reward function:
\begin{equation}
r(x, y_0) = \beta\, \mathbb{E}_{p_\theta(y_{1:T} \mid y_0, x)} \left[ \log \frac{p_\theta^*(y_{0:T} \mid x)}{p_{\text{ref}}(y_{0:T} \mid x)} \right] + \beta \log Z(x).
\end{equation}

Substituting this into the Listwise Preference Optimization (LiPO) framework, we obtain the Listwise preference losses:
\begin{multline}
\label{eq:lipodiff}
% \small
L_{\text{LiPO-Flow}}(\theta) = - \mathbb{E}_{x, y_1, \ldots, y_K \sim \mathcal{D}}  \\
\sum_{k=1}^K \log
\frac{
\exp \left( \mathbb{E}_{y^{\tau(k)}_{1:T}} \left[\beta \log \frac{\pi^*(y^{\tau(k)}_{0:T} \mid x)}{\pi_{\text{ref}}(y^{\tau(k)}_{0:T} \mid x)} \right] \right)
}{
\sum_{j=k}^K \exp \left( \mathbb{E}_{y^{\tau(j)}_{1:T}} \left[\beta \log \frac{\pi^*(y^{\tau(j)}_{0:T} \mid x)}{\pi_{\text{ref}}(y^{\tau(j)}_{0:T} \mid x)} \right] \right)
},
\end{multline}
where $y^{\tau(k)}_{1:T} \sim p_\theta(y_{1:T} \mid y^{\tau(k)}_0)$.

\begin{figure*}[th]
    \centering
    \includegraphics[width=\linewidth]{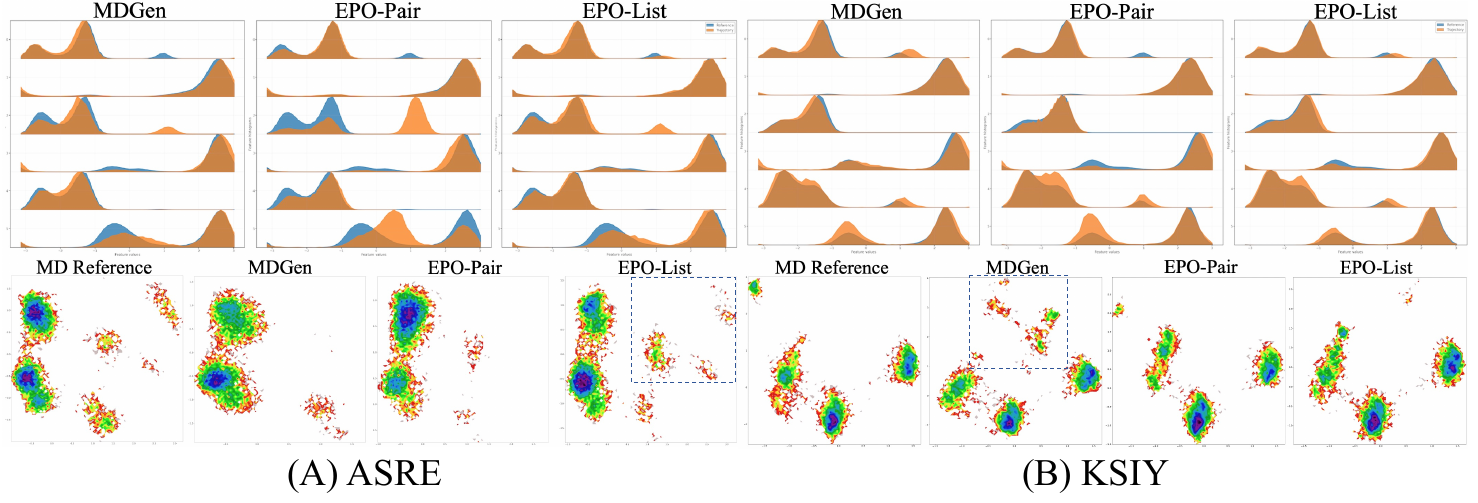}
    \caption{Comparison of torsion angle distributions and free energy surfaces (FES) for two tetrapeptide sequences. EPO successfully captures crucial metastable states absent in pretrained model outputs and effectively corrects high-energy biases observed in pretrained distributions (highlighted in dotted boxes, respectively), demonstrating strong alignment with the ground-truth energy landscapes.}
    \label{fig:tetra-torsion-fes}
\end{figure*}

\subsection{Practical Upper Bound}
The listwise loss derived above is intractable to optimize directly for two primary reasons.
First, the functional form of the Plackett-Luce choice probability is difficult to handle.
EPO addresses this by leveraging the function's convexity; applying Jensen's inequality yields a tractable upper bound.
Second, sampling from the reverse-time transition distribution, $p_\theta(y_{t-1,t}|y_0)$, is infeasible during training.
Following established practice for continuous-time generative models~\cite{flowdpo}, we approximate this intractable reverse transition with the corresponding forward process probability.

By combining these two solutions—the upper bound on the choice probability and the forward process approximation—and further simplifying a key resulting term with a Mean Squared Error (MSE) proxy between the predicted and reference trajectories, the final, practical objective is arrived:
\begin{multline}
L_{\text{LiPO-Flow Matching}}(\theta) \\ 
\leq  - \mathbb{E}_{(x,\mathcal{Y}), t, y^k_{t^*}} \sum_{k=1}^K \log \frac{\exp \big( \beta \log \frac{\pi^*(y_{T-1}^{\tau(k)} | y_T^{\tau(k)})}{\pi_{\text{ref}}(y_{T-1}^{\tau(k)} | y_T^{\tau(k)})} \big)}{\sum_{j=k}^K \exp \big( \beta \log \frac{\pi^*(y_{T-1}^{\tau(j)} \mid y_T^{\tau(j)})}{\pi_{\text{ref}}(y_{T-1}^{\tau(j)} \mid y_T^{\tau(j)})}} \big) \\
= - \mathbb{E}_{(x,\mathbf{Y}), t, y^k_{t^*}} \sum_{k=1}^K \log \frac{\exp \big[ s_\theta(\tau(k)) \big]}{\sum_{j=k}^K \exp \Big[ s_\theta(\tau(j)) \Big]},
\end{multline}
\begin{multline}
s_\theta(\tau(i)) \triangleq \beta \Big( \text{MSE}_t(y_0^{\tau(i)}, y_1^{\tau(i)}; \theta_{\text{ref}})  \\
 - \text{MSE}_t(y_0^{\tau(i)}, y_1^{\tau(i)}; \theta_{\text{opt}}) \Big),
\end{multline}
where 
$(x,\mathbf{Y}) \triangleq (x, y_1, \ldots, y_K) \sim \mathcal{D},$
$t \sim \mathcal{U}(0, T),$ 
$y^k_{t^*} \triangleq y_{t-1,t}^{k} \sim p_\theta(y_{t-1,t} \mid y_0^k), \, \forall k \in \{1, \ldots, K\}.$ 

\subsection{Ensemble Sampling as Forward Simulation}
\label{sec:forward_simulation_mdgen}
EPO uses MDGen~\cite{mdgen} as the backbone model to perform forward simulation of molecular trajectories $\boldsymbol{\chi} = [\mathbf{X}_1, \ldots, \mathbf{X}_T]$ for a given L-residue amino acid sequence $A$.
Each $\mathbf{X}_t \in \mathbb{R}^{3N}$ represents all-atom coordinates. MDGen processes proteins by defining local reference frames for each residue \citep{yim2023se}, and parameterizes molecular structures using per-residue $\mathrm{SE}(3)$ roto-translations $(R, \mathbf{t})$ and seven internal torsion angles $(\psi, \phi, \omega, \chi_1, \ldots, \chi_4)$, leading to
$    \boldsymbol{\chi}_t^l = ((R, \mathbf{t})_t^l, (\psi, \phi, \omega, \chi_1, \ldots, \chi_4)_t^l  \in \left( \left[ \mathrm{SE}(3) \times \mathbb{T}^7 \right]^L\right)^T$.
MDGen addresses the computationally prohibitive updates in each frame by transforming the absolute poses of the trajectories into a key-frame-based relative representation.
Given $K$ key frames ($K=1$ indicates the first trajectory frame in forward simulation, which is the setting in this paper),
the state of residue $j$ at frame $t$ (defined by its absolute roto-translation $g_t^j$ and torsions $\boldsymbol{\tau}_t^j$) is featurized into a token $\boldsymbol{\xi}_t^j$.
This token comprises $K$ relative roto-translations and the residue's torsions, embedded into a fixed-dimension vector:
\begin{multline}    
    \boldsymbol{\xi}^j_t = \Big([g_{t}^{j}]^{-1}g_t^j,\boldsymbol{\tau}_t^j\Big) \in SE(3) \times \mathbb{T}^7 \\
    \equiv (\hat{\mathbb{Q}}^+ \oplus \mathbb{R}^3)^1 \times (S^2)^7 \subset \mathbb{R}^{7+14}.
\end{multline}
A transformer-based model architecture guided by SiT \cite{ma2024sit} is used in MDGen, parameterizing a final velocity network $v_\theta$: $v_\theta(\cdot, t \mid \{g_{t_k}\}_{k=1}^K, A): \mathbb{R}^{T \times L \times (7K + 14)} \times[0,1] \rightarrow \mathbb{R}^{T \times L \times (7K + 14)}$.

\begin{table*}[t]
\centering
\begin{small}
\begin{tabular}{l|cccccccc}
\toprule
& \begin{tabular}{@{}c@{}}Pairwise\\RMSD $r$\\($\uparrow$)\end{tabular} & \begin{tabular}{@{}c@{}}Global\\RMSF $r$\\($\uparrow$)\end{tabular} & \begin{tabular}{@{}c@{}}Per-target\\RMSF $r$\\($\uparrow$)\end{tabular} & \begin{tabular}{@{}c@{}}Root\\$\mathcal{W}_2$\\($\downarrow$)\end{tabular} & \begin{tabular}{@{}c@{}}MD PCA\\$\mathcal{W}_2$\\($\downarrow$)\end{tabular} & \begin{tabular}{@{}c@{}}\%PC-sim$>$0.5\\($\uparrow$)\end{tabular} & \begin{tabular}{@{}c@{}}Weak\\contacts $J$\\($\uparrow$)\end{tabular} & \begin{tabular}{@{}c@{}}Exposed\\residue $J$\\($\uparrow$)\end{tabular} \\
\midrule
% AlphaFlow & 0.48/0.48 & 0.60/0.60 & 0.85/0.81 & 2.61/4.05 & 1.52/2.14 & 44 & 0.62 & 0.41 \\
MDGen & 0.48/0.42 & 0.50/0.49 & 0.71/0.70 & 2.69/3.47 & 1.89/2.43 & 10 & 0.51 & 0.29 \\
EPO-Pair & 0.49/0.49 & 0.49/0.49 & 0.74/0.69 & 2.88/3.53 & 1.88/2.42 & 9 & 0.42 & 0.43 \\
EPO-List & 0.51/0.51 & 0.50/0.50 & 0.75/0.70 & 2.79/3.43 & 1.77/2.35 & 13 & 0.45 & 0.44 \\
\bottomrule
\end{tabular}
\end{small}
\caption{Comparison of generated ensembles against ground-truth MD simulations on the ATLAS benchmark. Values reported are the median, with mean values shown after a backslash where applicable.}
\label{tab:atlas-distribution}
\end{table*}

\begin{table*}[bt]
\begin{center}
\begin{small}
% \begin{sc}
\begin{tabular}{lccccccc}
\toprule
\multirow{2}{*}{Models} & \multicolumn{4}{c}{JSD ($\downarrow$)} &  RMSE$_\text{contact}$  ($\downarrow$)   & RMSF   \\ 
\cmidrule(lr){2-5} & PwD &  Rg & TIC & TIC-2D & (Å)  &   (Å)  \\
\midrule
% Ref (N=1000) & 0.10/0.09 & 0.09/0.09 & 0.14/0.11 & 0.28/0.27 & 1.00/1.00 & 0.77/0.64 & 7.9/8.6 \\
% \midrule
EigenFold   & 0.53/0.56 & 0.52/0.55 & 0.50/0.50 & 0.64/0.66 & 6.18/6.22 & 1.6/1.1 \\
Str2Str-SDE & 0.34/0.32 & 0.30/0.24 & 0.39/0.38 & 0.56/0.58 & 3.68/4.01 & 7.8/8.0 \\
Str2Str-ODE & 0.37/0.38 & 0.33/0.30 & 0.40/0.39 & 0.57/0.59 & 4.14/4.36 & 6.4/6.3 \\
% \midrule
% \rowcolor{Highlight}
% ConfDiff-Base   & \textbf{0.29/0.27} & \textbf{0.25/0.22} & \textbf{0.36/0.37} & \textbf{0.52/0.52} & 0.89/0.91 & \underline{3.61/3.57} & 6.1/5.9 \\
% \rowcolor{Highlight}
ConfDiff-Energy  & 0.34/0.34 & 0.31/0.29 & 0.39/0.40 & 0.54/0.56  & 3.65/3.80 & 7.1/6.1   \\
% \rowcolor{Highlight}
ConfDiff-Force  & \underline{0.29/0.27} & \textbf{0.26/0.24} & \underline{0.38/0.38} & \underline{0.54/0.54} & {3.25/3.38} & 6.2/5.7 \\
\midrule
EPO-Pair & 0.30/0.30 & 0.29/0.29 & 0.39/0.40 & 0.57/0.61 & \underline{2.68/2.54} & 6.6/6.6 \\
EPO-List & \textbf{0.28/0.26} & \underline{0.28/0.26} & \textbf{0.32/0.30} & \textbf{0.49/0.47} & \textbf{2.36/2.28} &  6.3/6.6 \\ 
\bottomrule
\end{tabular}
\caption{Results on Fast-Folding proteins. All values are shown as mean/median.}
\label{tab:fastfolding}
\end{small}
\end{center}
% \vspace{-4mm}
\end{table*}

\section{Experiments}
\label{sec:experiments}
In this section, we conduct comprehensive experiments to rigorously evaluate the effectiveness of the proposed EPO framework.
Our main findings can be summarized as follows:
(1) EPO establishes new state-of-the-art performance across nine evaluation metrics on three widely adopted protein conformation benchmarks.
(2) By relying solely on energy-based labels, EPO promotes more effective exploration of the conformational space and improves alignment of the generated ensembles with the Boltzmann energy distribution.
(3) Despite theoretical equivalence, the listwise approach outperforms its pairwise counterpart in practical experiments.

\subsection{Implementation Details}
\paragraph{Datasets.} 
The adopted benchmarks span a range of molecular systems—from small peptides to large, structurally varied proteins and fast-folding systems—allowing us to assess our model's capabilities across different scales and complexities.
Specifically, we first evaluate the Tetrapeptides~\cite{mdgen} dataset, which contains all-atom molecular dynamics (MD) trajectories for $3,000$ training, $100$ validation, and $100$ test tetrapeptides, each simulated for $100\,\text{ns}$.
We then conduct finetuning on ATLAS~\cite{atlas} dataset, which is constructed to maximize domain-level structural coverage across ECOD X-classes.
It consists of explicit-solvent, all-atom MD simulations, providing three independent $100\,\text{ns}$ trajectories for each of its $1,390$ structurally diverse protein entries.
We also evaluate the Fast-Folding benchmark~\cite{fastfolding}, which includes MD simulation data for $12$ small proteins exhibiting rapid folding-unfolding dynamics.

We follow the standard dataset splits as in MDGen~\cite{mdgen} and AlphaFlow~\cite{alphaflow} for the three datasets.
\paragraph{Baselines.}
EigenFold~\cite{eigenfold} is the first to leverage diffusion models for this task, generating conformational ensembles by sampling from a learned distribution over protein eigenmodes. 
MDGen~\cite{mdgen} directly models MD trajectories, utilizing a Scalable Interpolant Transformer and key-frame conditioning for improved computational efficiency. 
Str2Str~\cite{str2str} uses online stochastic perturbations during generation to enhance diversity, which is similar to EPO, but it does not incorporate explicit physical guidance from energy or force signals.
ConfDiff~\cite{confdiff} are fine-tuned on static, offline datasets annotated with pre-computed energy and force labels, which limits their ability to dynamically adapt to distributional shifts during training. 
A more detailed discussion of these related works is provided in the Appendix.

\begin{table}[t]
\small
\centering
\begin{tabular}{l|ccccc}
\toprule
% & Torsions (bb) & Torsions (sc) & Torsions (all) & TICA-0 & TICA-0,1 joint \\
& $\chi_{\text{bb}}$ & $\chi_{\text{sc}}$ & $\chi_{\text{all}}$ & TICA-0 & TICA-0,1  \\
\midrule
Reference & 0.103 & 0.055 & 0.076 & 0.201 & 0.268\\
\midrule
MDGen & 0.130 & \textbf{0.093} & 0.109 & 0.230 & 0.316 \\
EPO-Pair & 0.127 & 0.098 & 0.110 & 0.237 &0.318 \\
EPO-List & \textbf{0.125} & \textbf{0.093} & \textbf{0.107} & \textbf{0.226} & \textbf{0.311} \\
\bottomrule
\end{tabular}
\caption{Jensen-Shannon divergence between ground-truth and generated tetrapeptide distributions. Lower is better.}
% \vspace{-4mm}
\label{tab:tetrapeptides-jsd}
\end{table}

\paragraph{Experimental Settings} 
The reference policy model, $\pi_{ref}$, is based on MDGen checkpoints separately pretrained on the Tetrapeptides and ATLAS datasets.
For the Tetrapeptides experiments, $\beta$ is set to 1 and the score norm of SDE sampling is 0.01; while for the ATLAS experiments, $\beta$ is set to 250 and the score norm is 0.0001.
The three repeats for each ATLAS protein are merged with equal weight and randomly sampled during training.
The denoising step is both set to 50, and the learning rate is both set to 1e-5.
We use grid search to select the above hyperparameters, and the sensitivity analysis can be found in ablation studies.
We adopt Madrax~\cite{orlando2024integrating}, a differentiable empirical force field implemented in PyTorch, as the physics-based energy reward. 
Low-Rank Adaptation (LoRA)~\cite{hu2022lora} is employed for parameter-efficient fine-tuning of pretrained models.

Training the EPO model on the Tetrapeptides dataset requires 30 hours, while the ATLAS dataset requires 72 hours.
This process utilizes four A100 40GB GPUs on a Linux 5.4.0 platform with PyTorch 1.12 and CUDA 11.3.
All reported metrics are from a single experimental run, a protocol consistent with the baseline models we compare against.

\subsection{Tetrapeptides}
Table~\ref{tab:tetrapeptides-jsd} assesses the Jensen-Shannon divergence (JSD) between the generated and ground-truth trajectories across two sets of collective variables:
(1) the individual backbone and sidechain torsion angles ($\chi_{\text{bb}}$, $\chi_{\text{sc}}$, $\chi_{\text{all}}$) for each tetrapeptide; 
(2) the leading independent components extracted via time-lagged independent component analysis (TICA), capturing the slowest dynamical modes of peptides. 
EPO consistently achieves superior distributional alignment with the reference MD trajectories, closely matching the accuracy obtained by replicate 100-ns simulations.

We further visualize the torsion angle distributions and the free energy surface (FES) along the two primary TICA components in Figure~\ref{fig:tetra-torsion-fes}. 
The torsion angle distribution plots clearly demonstrate that EPO-generated trajectories (blue) closely approximate the ground-truth MD trajectories (orange). 
Notably, the FES analysis reveals that EPO not only induces novel metastable states absent in the pretrained baseline (as in sequence ASRE) but also effectively corrects high-energy biases present in the pretrained model (as in sequence KSIY). 
% Importantly, these refinements maintain a high overall similarity to the ground-truth energy landscapes. 
% We attribute these improvements to EPO's capability to efficiently overcome energy barriers through online sampling, as well as the effectiveness of energy-based preference optimization.
These results validate our central hypothesis: EPO’s online sampling framework, combined with energy-based preference optimization, provides an effective and reliable mechanism for crossing critical energy barriers, thus enabling the accurate recovery of physically realistic conformational states.

\begin{figure}[t]
    \centering
    \includegraphics[width=\columnwidth]{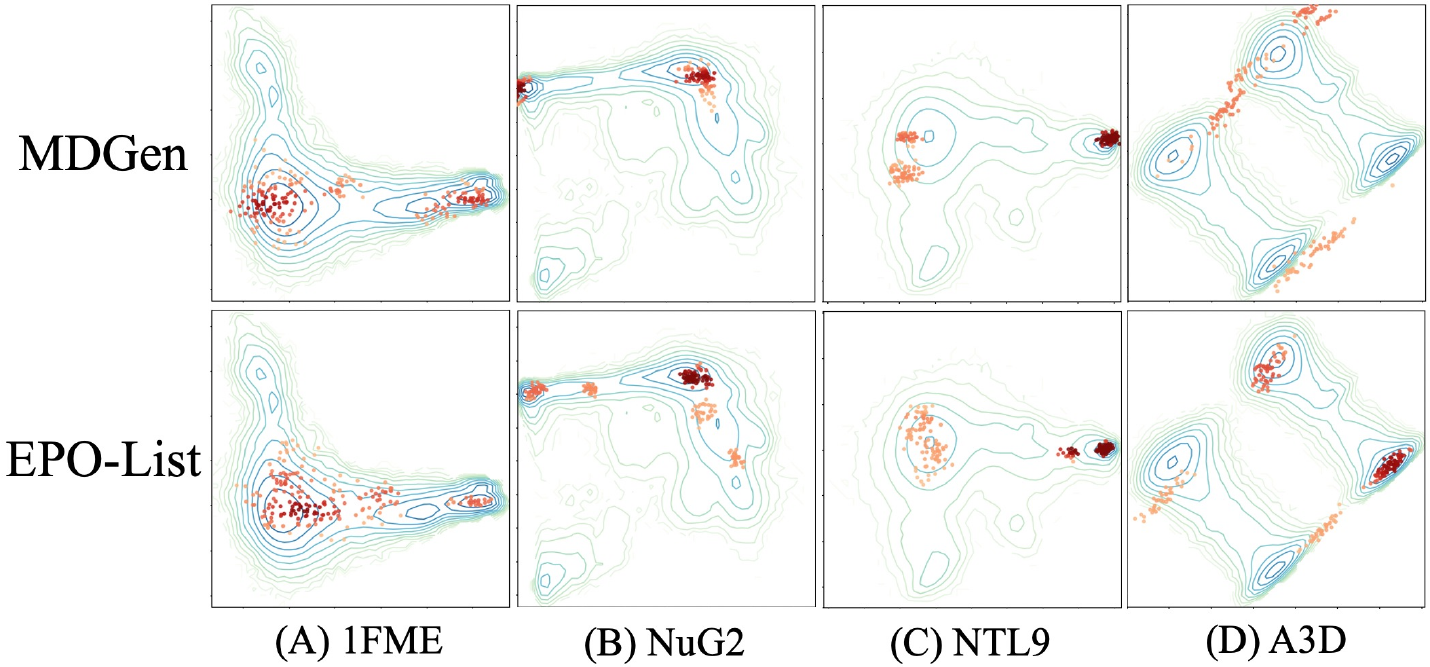}
    \caption{Sample distributions projected onto the first two time-lagged independent components for four proteins from the Fast-Folding dataset. EPO-List demonstrates improved diversity by exploring a broader conformational landscape. }
    \label{fig:Fast-Folding-fes}
% \vspace{-4mm}
\end{figure}

\begin{figure*}[t]
    \includegraphics[width=\textwidth]{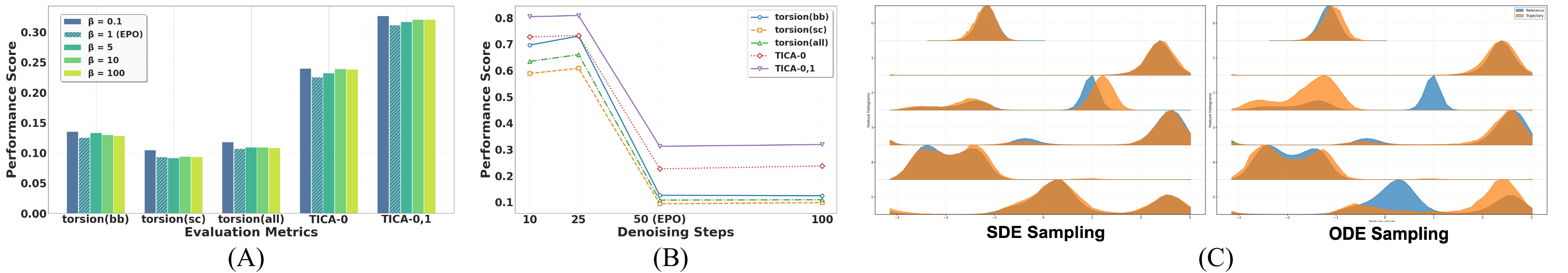} 
    \caption{Analysis of key components of EPO. (A) Impact of the temperature hyperparameter $\beta$. (B) Effect of the number of denoising steps in SDE sampling. An insufficient number of steps causes training instability, whereas an excessive number provides no significant improvement. (C) Torsion angle distributions for SDE vs. ODE strategies on the SPFH sequence. The SDE-based model successfully identifies distant peaks corresponding to metastable states separated by high energy barriers.}
    \label{fig:ablation}
% \vspace{-2mm}
\end{figure*}

\subsection{ATLAS}
Next, we evaluate EPO on the 82 test proteins in ATLAS.
We report the mean and median values of the distributional metrics to quantify discrepancies between the generated and reference MD ensembles.
For detailed definitions and descriptions of the evaluation metrics, please refer to Appendix.
As shown in Table~\ref{tab:atlas-distribution}, EPO exhibits advantages over baseline methods in several key distributional metrics, including higher correlations in pairwise RMSD and per-target RMSF, as well as lower PCA $\mathcal{W}_2$ distances. 

We attribute the modest performance gains on the ATLAS benchmark primarily to the small score norm used in SDE sampling for ATLAS proteins.
Larger score norms caused sampling instability and structural collapse, making this adjustment necessary when applying SDE to longer protein chains.
However, the reduced stochastic exploration likely limited the model's ability to discover a more diverse and accurate conformational ensemble, underscoring an important direction for future research.

\subsection{Fast-Folding}
We further evaluate the transferability and generalization of EPO by performing zero-shot testing on the Fast-Folding protein benchmark. 
Specifically, we directly apply the checkpoint optimized on the ATLAS dataset to generate ensembles without any fine-tuning or adaptation.

Table~\ref{tab:fastfolding} summarizes the performance of EPO-generated conformational ensembles in comparison to several established baseline methods.
We employ the evaluation metrics introduced in ConfDiff~\cite{confdiff}, where detailed explanations can be found in Appendix.

EPO establishes four new state-of-the-art results and achieves second-best performance on the remaining distributional metrics.
Notably, comparing with closely related baselines, i.e., Str2Str and ConfDiff, highlights the core advantages of EPO.
Unlike Str2Str, EPO's online exploration is effectively steered by an explicit physical signal. 
And unlike ConfDiff, EPO's physical guidance is applied dynamically in real-time, avoiding the constraints and biases of training on a static, offline dataset. 
This synergy of online guidance and online exploration is what allows EPO to continuously discover novel and physically favorable states.

\begin{table}[t]
\small
\centering
\begin{tabular}{l|ccccc}
\toprule
% & Torsions (bb) & Torsions (sc) & Torsions (all) & TICA-0 & TICA-0,1 joint \\
& $\chi_{\text{bb}}$ & $\chi_{\text{sc}}$ & $\chi_{\text{all}}$ & TICA-0 & TICA-0,1  \\
\midrule
EPO-ODE & {0.167} & {0.107} & {0.132} &{0.247} & {0.337} \\
EPO-SDE & {0.125} & {0.093} & {0.107} &{0.226} & {0.311} \\
\bottomrule
\end{tabular}
\caption{Comparison between SDE and ODE sampling strategy on tetrapeptides experiments. Lower is better.}
% \vspace{-4mm}
\label{tab:ablation-tetra-sdeode}
\end{table}

\subsection{Ablation Studies}

In this section, we investigate and validate EPO’s sensitivity to key components.

First, we examine the temperature hyperparameter $\beta$, critical for scaling the preference objective. As shown in Figure~\ref{fig:ablation}(A), setting $\beta$ to a very small value is detrimental and can be counterproductive.
This highlights that $\beta$'s optimal value is context-dependent and must be carefully selected.

Next, we analyze the effect of denoising steps on SDE sampling. 
Figure~\ref{fig:ablation}(B) shows that insufficient steps cause unstable training and mode collapse, as the reverse trajectory fails to generate valid conformations.
Conversely, increasing the number of steps beyond a certain point yields diminishing returns; it significantly increases computational cost without a corresponding improvement in sample quality or model performance.
This highlights the need for a balance between stability and tractability.

Finally, we compare the exploratory power of stochastic (SDE) versus deterministic (ODE) sampling.
Figure~\ref{fig:ablation}(C) presents torsion angle distributions on the SPFH sequence for both.
The SDE approach shows a significant advantage, identifying distant peaks corresponding to distinct metastable states separated by high energy barriers. 
In contrast, the ODE sampler converges to limited local minima, failing to discover these crucial conformations. 
More distributional visualizations can be found in the Appendix.

\section{Conclusions and Limitations}
In this work, we have presented Energy Preference Optimization (EPO), an online refinement framework that aligns pretrained protein-conformation generators with the Boltzmann distribution using direct physics-based feedback. 
By combining an online SDE sampling strategy with a listwise preference optimization, EPO successfully produces diverse and physically realistic ensembles while eliminating the need for additional molecular dynamics simulations.

% However, two primary limitations motivate important directions for future research. 
% First, the online refinement loop is computationally intensive due to its per-sample energy evaluations.
% This necessitates the development of more efficient refinement protocols; future work will explore coarse-grained force fields and models to reduce this cost while maintaining fidelity at relevant resolutions.
% Second, while our listwise preference objective demonstrates strong empirical performance, it currently lacks theoretical guarantees of convergence to the target Boltzmann distribution, motivating subsequent investigation into optimization frameworks that offer provable convergence properties for physical systems. 
% Looking ahead, we also plan to scale EPO to more expressive generative models with stronger capacity for large proteins and complex systems, leveraging recent advances in RLHF-style frameworks and sampling strategies developed for large language models.

Two primary limitations motivate future research. 
First, the computationally intensive online refinement loop necessitates more efficient protocols.
Coarse-grained models may reduce this cost while maintain fidelity. 
Second, our listwise preference objective, while empirically strong, lacks theoretical guarantees of convergence to the target Boltzmann distribution, motivating investigation into optimization frameworks with provable convergence. 
Looking ahead, we plan to scale EPO to more expressive generative models for large proteins, and leverage engineering technologies in recent RLHF-style frameworks.

\section*{Acknowledgement}
This work is supported by the Administrative Committee of Zhongguancun Science City. This work is also supported by Beijing Natural Science Foundation (L243006).

\appendix
\section{Related Work}
\subsection{Deep Learning Generative Models for Protein Ensemble Generation}
\paragraph{Heuristic Extension of Single-state Predictors.}
Initial computational approaches for generating protein conformational ensembles often relied upon powerful single-state structure prediction methods, such as AlphaFold~\citep{jumper2021highly} and RoseTTAFold~\citep{baek2021accurate}. 
These approaches generated ensembles heuristically, typically by varying input conditions—such as sampling diverse Multiple Sequence Alignments (MSAs)~\citep{af3,schafer2025sequence}—or by clustering outputs from repeated predictions under perturbed conditions~\citep{afcluster,kalakoti2025afsample2}.
Although these heuristic strategies revealed some conformational variability, they lacked explicit probabilistic modeling of the underlying conformational distributions and were inherently limited in capturing rare or transient states essential for protein function~\citep{karplus2005molecular,henzler2007dynamic}.

\paragraph{Distribution Learning from MD Simulation.}
To explicitly model protein conformational distributions, recent research has increasingly adopted deep generative frameworks, primarily leveraging SE(3)-equivariant diffusion models and normalizing flow techniques. 
To align generated ensembles with physically realistic Boltzmann distributions, several approaches have introduced explicit energy guidance, commonly achieved by fine-tuning pretrained generative models with molecular dynamics (MD) simulation data or by incorporating experimentally derived energy potentials during the sampling process~\citep{mdgen, alphaflow, confdiff, bioemu, klein2023timewarp, schreiner2023implicit}.
Recent works like EBA~\cite{lu2025eba} and P2DFlow~\cite{jin2025p2dflow} explicitly incorporate physical energy or Boltzmann-derived factors into optimization, but face the challenges of numerical instability due to the large variance in energy values across conformational landscape.
While these MD-based refinement methods have substantially improved the physical plausibility of generated conformations, their dependence on simulated trajectories inherently introduces limitations, such as finite timescale coverage, inaccuracies in force fields, and sampling biases. 
Consequently, these constraints restrict models from effectively exploring novel, biologically relevant conformational states absent from their training sets.
An alternative research direction integrates physically informed experimental data—such as Nuclear Magnetic Resonance (NMR) or Cryo-Electron Microscopy (Cryo-EM)—to enhance the diversity and realism of conformational samples produced by generative models~\citep{laurents2022alphafold, maddipatla2025inverseproblemsexperimentguidedalphafold, tengcryogen, renphysical, gyawali2025multimodal}.

\subsection{Reinforcement Learning from Human Feedback}
Aligning large language models (LLMs) with user intent was first operationalized through Reinforcement Learning from Human Feedback (RLHF) \citep{christiano2017deep, stiennon2020learning}.  
In the canonical pipeline, pairwise preference data are used to train a reward model, and the policy is subsequently improved with on-policy algorithms such as PPO \citep{ppo}.  
Despite impressive empirical success, RLHF inherits the sample inefficiency and brittle hyperparameter tuning of reinforcement learning.  
% Several attempts have been made to mitigate these issues without abandoning the RL formalism entirely.  
% Offline variants such as Implicit Q-Learning (IQL) and Advantage-Weighted Regression (AWR) \citep{peng2019awr, kostrikov2021offline, qin2024align} reweight logged trajectories by estimated advantages, thereby avoiding expensive online roll-outs.  
% Constitutional AI \citep{bai2024constitutional} replaces external preference labels with self-critique signals and thus removes part of the human-in-the-loop cost, but still relies on an RLHF-style outer loop.  
% Collectively, these approaches highlight both the versatility and the lingering complexity of RL-centric alignment.

To bypass reward modelling and online exploration altogether, \citeauthor{dpo} proposed Direct Preference Optimization (DPO).  
DPO reframes preference alignment as a purely offline, KL-regularized classification problem whose gradients depend only on logged comparisons, enabling stable training with standard supervised infrastructure.  
Building on the same “likelihood-based” philosophy, several variants have been introduced.  
Most recently, \citeauthor{ethayarajh2024kto} draw on Kahneman--Tversky prospect theory to design KTO, a human-aware loss that directly maximizes subjective utility rather than log-likelihood.
In parallel, \citeauthor{wang2023beyond} present $f$-DPO, which replaces the reverse-KL term in DPO with a family of tractable $f$-divergence constraints (e.g., Jensen–Shannon, forward-KL, $\alpha$-divergences), yielding a supervised objective that balances alignment and diversity and outperforms PPO in divergence efficiency.

Although the original formulation targets discrete text generation, the underlying principle—matching the Boltzmann rational distribution implied by preferences—extends naturally to continuous data.  
\citeauthor{wallace2024diffusion,yang2024using} instantiate this idea in computer vision with Diffusion-DPO, combining a score-based diffusion backbone with a DPO-style loss.
While pairwise comparisons are convenient, many real-world applications supply graded or ranked feedback.
ListDPO~\citep{lipo} augments the DPO objective with list-wise information through a $\lambda$-rank weighting scheme.  
The resulting loss directly optimizes the target ranking metric Normalized Discounted Cumulative Gain (NDCG), yielding consistent gains on retrieval-augmented generation and recommendation tasks.
\citeauthor{gu2024aligning, wang2024training} utilize external rewards for diffusion models but focus on specific properties.
Other methods employ RL~\cite{black2023training, fan2023dpok} or rely on direct backpropagation via a differentiable reward signal~\cite{clark2023directly}.
To the best of our knowledge, we are the first to extend this post-training paradigm to the generation of protein ensembles.

% \section{Alogrithm}
% \begin{minipage}{\linewidth}
% \begin{algorithm}[H]
% \caption{Energy Preference Optimization (with flow matching)}
% \begin{algorithmic}[1]
% \STATE \textbf{Input:} Pre-trained model $\theta_{\text{ref}}$, energy function $E$, gradient threshold $\text{grad\_thre}$
% \STATE Initialize $\theta_{\text{opt}}$ with $\theta_{\text{ref}}$ and adapt LoRA, freeze $\theta_{\text{ref}}$
% \WHILE{not converged}
% \STATE Sample $x_0 \sim \mathcal{N}(0, 1)$ \COMMENT{improve efficiency}
% \STATE $x_1 = \text{sample\_odeint}(\theta_{\text{ref}}(x_t, t))$, integrating $t$ from $0$ to $1$, where \textbf{if}{$t < \text{grad\_thre}$} Apply $\text{torch.no\_grad()}$ to $\theta_{\text{ref}}$
% \STATE Compute energy $E = E(x_1)$
% \STATE Calculate ranking $\tau$ based on energy values
% \STATE Sample $s \sim \mathcal{U}(t, 1)$
% \STATE Calculate flow matching loss $\text{MSE}(x_0, x_1, \theta)$ accroding to Equation5.
% \STATE Compute final loss $l$ using energy preferences and flow matching loss according to Equation8
% \STATE Update $\theta_{\text{opt}}$ with $l$
% \ENDWHILE
% \RETURN $\theta_{\text{opt}}$
% \end{algorithmic}
% \end{algorithm}
% \end{minipage}

\begin{figure*}
    \centering
    \includegraphics[height=\textheight]{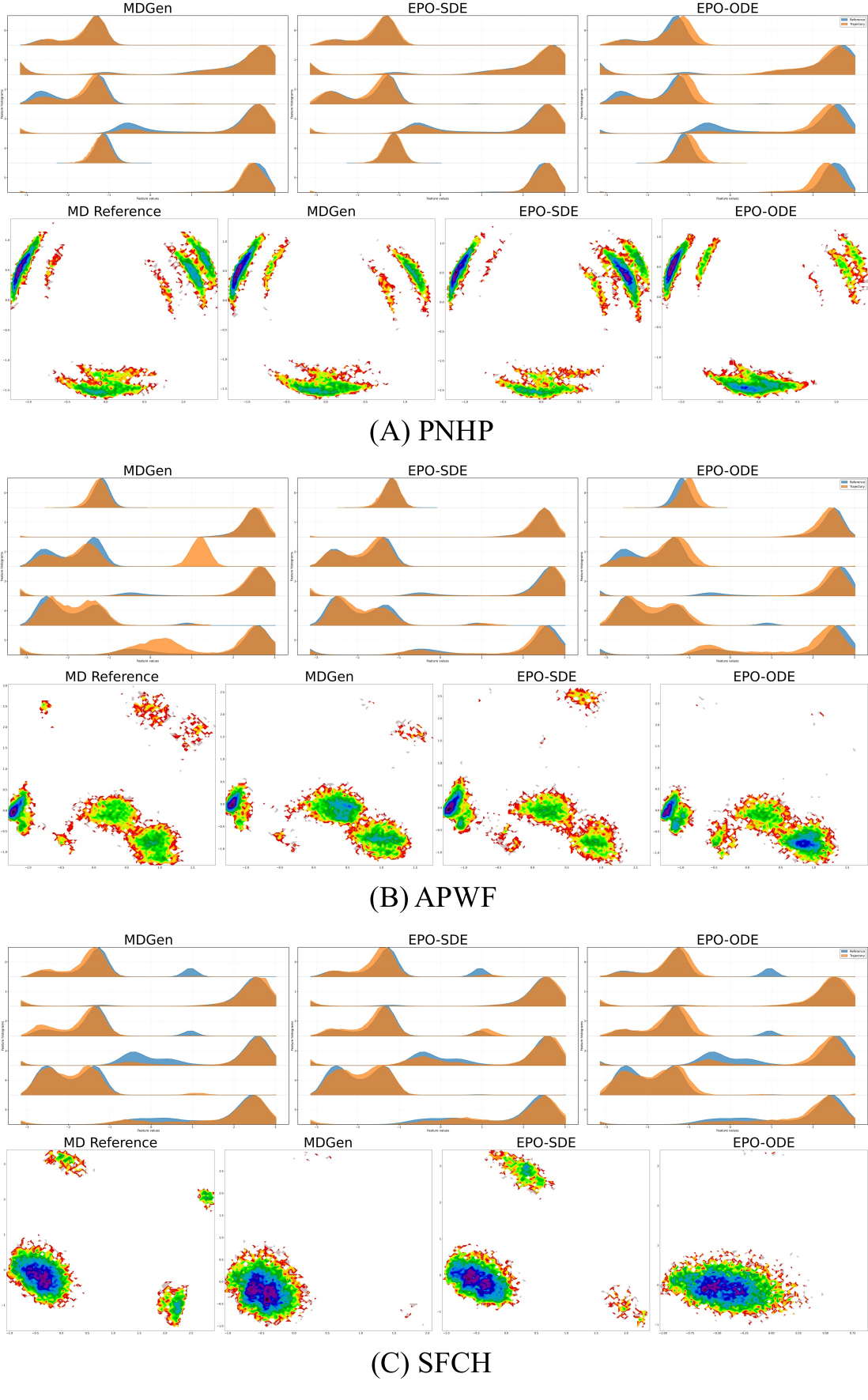}
    \caption{Torsion angle distribution and free energy surface comparison between SDE and ODE strategies for Tetrapeptides experiment.}
    \label{fig:ablation_sdeode_appendix}
\end{figure*}

\section{Details of the EPO Derivation}
\subsection{From DPO to LiPO}
The LiPO approach utilizes the Plackett-Luce model to handle preference rankings over multiple outputs, generalizing pairwise preference methods like the Bradley-Terry model. Similar to the Bradley-Terry model, it stipulates that when presented with a set of possible choices, people prefer a choice with probability proportional to the value of some latent reward function for that choice. It can be formulated as follows:
\begin{equation}
p^*(\tau \mid y_1, \ldots, y_K, x) = \prod_{k=1}^K \frac{\exp\left(r^*(x, y_{\tau(k)})\right)}{\sum_{j=k}^K \exp\left(r^*(x, y_{\tau(j)})\right)}
\end{equation}
In the LiPO framework for language models, the reward function is defined over the generated output \( y \) given an input \( x \). It mirrors the Direct Preference Optimization (DPO) reward under KL constraints and is expressed as:
\begin{equation}
    r(x, y) = \beta \log \frac{\pi_r(y \mid x)}{\pi_{\text{ref}}(y \mid x)} + \beta \log Z(x)
\end{equation}
Substituting the reward function in the PL model and utilizing the maximum likelihood, we can obtain the original LiPO objective:
\begin{multline}
    \mathcal{L}_{\text{DPO}}(\pi_\theta, \pi_{\text{ref}})
    = - \mathbb{E}_{\tau, y_1, \ldots, y_K, x \sim \mathcal{D}}  \\ \left[ 
\log \prod_{k=1}^K \frac{
    \exp\left( \beta \log \frac{\pi_\theta(y_{\tau(k)} \mid x)}{\pi_{\text{ref}}(y_{\tau(k)} \mid x)} \right)
}{
    \sum_{j=k}^K \exp\left( \beta \log \frac{\pi_\theta(y_{\tau(j)} \mid x)}{\pi_{\text{ref}}(y_{\tau(j)} \mid x)} \right)
}
\right]
\end{multline}

\subsection{Reward over the Time Path}
Same as diffusion DPO, we introduce latents \( y_{1:T} \) and define \( R(x, y_{0:T}) \) as the reward on the whole chain, such that we can define \( r(x, y) \) as
\begin{equation}
r(x, y) = \mathbb{E}_{\pi_0(y_{1:T}|y,x)} [R(x, y_{0:T})].
\end{equation}
Under KL constraints, the RLHF aims to optimize the distribution \(\pi_\theta(y|x)\) such that the reward model \(r(x, y)\) defined on it is maximized: 
\begin{multline}
\max_{\pi_\theta} \mathbb{E}_{x\sim \mathcal{D}_x,\, y \sim \pi_\theta(y \mid x)} \left[ r(\mathbf{x}, \mathbf{y}) \right] \\
- \beta \mathbb{D}_{\text{KL}} \left[ \pi_\theta(y \mid \mathbf{x}) \,\|\, \pi_{\text{ref}}(y \mid \mathbf{x}) \right]
\end{multline}
Inspired by diffusion DPO, we instead optimize its minimize its upper bound joint KL-divergence
\begin{align}
\min_{\pi_\theta}\;&
- \mathbb{E}_{\pi_\theta(y_0 \mid \mathbf{x})}
  \frac{r(\mathbf{x}, \mathbf{y}_0)}{\beta}\nonumber\\&
+ \mathbb{D}_{\text{KL}}\!\left(
   \pi_\theta(y_0 \mid \mathbf{x})
   \,\|\, 
   \pi_{\text{ref}}(y_0 \mid \mathbf{x})
\right)\\[6pt]
\leq\;&
\min_{\pi_\theta}\;
- \mathbb{E}_{\pi_\theta(y_0 \mid \mathbf{x})}
  \frac{r(\mathbf{x}, \mathbf{y}_0)}{\beta}\nonumber\\&
+ \mathbb{D}_{\text{KL}}\!\left(
   \pi_\theta(y_{0:T} \mid \mathbf{x})
   \,\|\, 
   \pi_{\text{ref}}(y_{0:T} \mid \mathbf{x})
\right)\nonumber\\[6pt]
=\;&
\min_{\pi_\theta}\;
- \mathbb{E}_{\pi_\theta(y_{0:T} \mid \mathbf{x})}
  \frac{R(\mathbf{x}, \mathbf{y}_{0:T})}{\beta}\nonumber\\&
+ \mathbb{D}_{\text{KL}}\!\left(
   \pi_\theta(y_{0:T} \mid \mathbf{x})
   \,\|\, 
   \pi_{\text{ref}}(y_{0:T} \mid \mathbf{x})
\right)\nonumber\\[6pt]
=\;&
\min_{\pi_\theta}\;
\mathbb{E}_{\pi_\theta(y_{0:T} \mid \mathbf{x})}\nonumber\\&
\!\Biggl(
\log \frac{\pi_\theta(y_{0:T} \mid \mathbf{x})}
{\pi_{\text{ref}}(y_{0:T} \mid \mathbf{x}) 
   \exp\!\left(\! R(\mathbf{x}, \mathbf{y}_{0:T})/\beta \!\right)/Z(\mathbf{x})}
- \log Z(\mathbf{x})
\Biggr)\nonumber\\[6pt]
=\;&
\min_{\pi_\theta}\;
\mathbb{D}_{\text{KL}}\!\nonumber\\&\left(
\pi_\theta(y_{0:T} \mid \mathbf{x})
\,\bigg\|\,
\pi_{\text{ref}}(y_{0:T} \mid \mathbf{x})
\frac{\exp\!\left(\! R(\mathbf{x}, \mathbf{y}_{0:T})/\beta \!\right)}
     {Z(\mathbf{x})}
\right)\!.\label{eq:KL_min}
\end{align}

The optimal \(\pi_\theta(y|x)\) of \eqref{eq:KL_min} has a unique closed-form solution:
\begin{equation}
\pi_\theta^*(\mathbf{y}_{0:T} \mid \mathbf{x}) = \frac{\pi_{\text{ref}}(\mathbf{y}_{0:T} \mid \mathbf{x}) \exp\left(R(\mathbf{x}, \mathbf{y}_{0:T})/\beta\right)}{Z(\mathbf{x})}
\label{eq:optimal-policy}
\end{equation}
Hence, we can obtain the original reward \(r(x,y)\) parameterized by probability path:
\begin{equation}
r(\mathbf{x}, \mathbf{y}_0) = \beta \, \mathbb{E}_{\pi_\theta(\mathbf{y}_{1:T} \mid \mathbf{y}_0, \mathbf{x})} \left[ \log \frac{\pi^*_\theta(\mathbf{y}_{0:T} \mid \mathbf{x})}{\pi_{\text{ref}}(\mathbf{y}_{0:T} \mid \mathbf{x})} \right] + \beta \log Z(\mathbf{x})
\label{eq:reward-from-kl}
\end{equation}

\subsection{LiPO Extension to Flow Matching Models}
Substituting this into the Listwise   Optimization (LiPO) framework, we obtain the Listwise preference losses:
\begin{multline}
\label{eq:lipodiff}
L_{\text{LiPO-Flow}}(\theta) = - \mathbb{E}_{x, y_1, \ldots, y_K \sim \mathcal{D}}  \sum_{k=1}^K \log \\
\frac{
\exp \left( \mathbb{E}_{y^{\tau(k)}_{1:T} \sim \pi_\theta(y_{1:T} \mid y^{\tau(k)}_0)} \left[\beta \log \frac{\pi^*(y^{\tau(k)}_{0:T} \mid x)}{\pi_{\text{ref}}(y^{\tau(k)}_{0:T} \mid x)} \right] \right)
}{
\sum_{j=k}^K \exp \left( \mathbb{E}_{y^{\tau(j)}_{1:T} \sim \pi_\theta(y_{1:T} \mid y^{\tau(j)}_0)} \left[\beta \log \frac{\pi^*(y^{\tau(j)}_{0:T} \mid x)}{\pi_{\text{ref}}(y^{\tau(j)}_{0:T} \mid x)} \right] \right)
}.
\end{multline}
We define \( f_\theta(\tau(k)) \) as follows:
\begin{align*}
&\beta T \mathbb{E}_t  
\mathbb{E}_{\mathbf{y}_{t-1} \sim q(\mathbf{y}_{t-1} \mid \mathbf{y}_t, \mathbf{y}^{\tau(k)}_0)}f_\theta(\tau(k))\\
&= \beta \mathbb{E}_{\mathbf{y}_{1, T} \sim q(\mathbf{y}_{1, T} \mid \mathbf{y}^{\tau(k)}_0)} \left[  \log \frac{\pi^*(y^{\tau(k)}_{0:T} \mid x)}{\pi_{\text{ref}}(y^{\tau(k)}_{0:T} \mid x)}\right]  \\
&=  \beta \mathbb{E}_{\mathbf{y}_{1, T} \sim q(\mathbf{y}_{1, T} \mid \mathbf{y}^{\tau(k)}_0)} \left[ \sum_{t=1}^{T} \left[  \log \frac{\pi^*(y_{t-1}^{\tau(k)} | y_{t}^{\tau(k)})}{\pi_{\text{ref}}(y_{t-1}^{\tau(k)} | y_{t}^{\tau(k)})} \right] \right]  \\
&=  \beta \mathbb{E}_{\mathbf{y}_{1, T} \sim q(\mathbf{y}_{1, T} \mid \mathbf{y}^{\tau(k)}_0)} T \mathbb{E}_t \left[  \log \frac{\pi^*(y_{t-1}^{\tau(k)} | y_{t}^{\tau(k)})}{\pi_{\text{ref}}(y_{t-1}^{\tau(k)} | y_{t}^{\tau(k)})} \right]  \\
&=  \beta T \mathbb{E}_t \mathbb{E}_{\mathbf{y}_{t-1}, t \sim q(\mathbf{y}_t \mid \mathbf{y}^{\tau(k)}_0)} \left[  \log \frac{\pi^*(y_{t-1}^{\tau(k)} | y_{t}^{\tau(k)})}{\pi_{\text{ref}}(y_{t-1}^{\tau(k)} | y_{t}^{\tau(k)})} \right]  \\
&=  \beta T \mathbb{E}_t  
\mathbb{E}_{\mathbf{y}_{t-1} \sim q(\mathbf{y}_{t-1} \mid \mathbf{y}_t, \mathbf{y}^{\tau(k)}_0)} \left[  \log \frac{\pi^*(y_{t-1}^{\tau(k)} | y_{t}^{\tau(k)})}{\pi_{\text{ref}}(y_{t-1}^{\tau(k)} | y_{t}^{\tau(k)})} \right]
\end{align*}
Replacing \( f(\theta) \) into the list preference losses:
\begin{multline}
\label{eq:lipodiff_replace}
L_{\text{LiPO-Flow}}(\theta) = - \mathbb{E}_{x, y_1, \ldots, y_K \sim \mathcal{D}} \sum_{k=1}^K \\ \log
\frac{
\exp \left( \beta T \mathbb{E}_t  
\mathbb{E}_{\mathbf{y}_{t-1} \sim q(\mathbf{y}_{t-1} \mid \mathbf{y}_t, \mathbf{y}^{\tau(k)}_0)}f_\theta(\tau(k)) \right)
}{
\sum_{j=k}^K \exp \left( \beta T \mathbb{E}_t  
\mathbb{E}_{\mathbf{y}_{t-1} \sim q(\mathbf{y}_{t-1} \mid \mathbf{y}_t, \mathbf{y}^{\tau(j)}_0)}f_\theta(\tau(j)) \right)
}.
\end{multline}
Since the denominator involves a log-exp-sum, which is convex, we can utilize Jensen's inequality:
\begin{multline}
\label{eq:lipodiff_final}
L_{\text{LiPO-Flow}}(\theta) \leq - \mathbb{E}_{x, y_1, \ldots, y_K \sim \mathcal{D},\, t \sim \mathcal{U}(0, T), 
\atop y_{t-1,t}^{1} \sim \pi_\theta(y_{t-1,t} \mid y_0^1),\, \ldots , y_{t-1,t}^{K} \sim \pi_\theta(y_{t-1,t} \mid y_0^K)}
\\ \sum_{k=1}^K \log
\frac{
\exp \left(\beta T f_\theta(\tau(k)) \right)
}{
\sum_{j=k}^K \exp \left(\beta T f_\theta(\tau(j)) \right)
}.
\end{multline}

Finally, we can substitute the flow formulation to obtain the desired objectives:
\begingroup
\small
\begin{multline}
L_{\text{LiPO-Flow Matching}}(\theta) \leq  \\
   - \mathbb{E}_{x, y_1, \ldots, y_K \sim \mathcal{D},\, t \sim \mathcal{U}(0, T), 
\atop y_{t-1,t}^{1} \sim \pi_\theta(y_{t-1,t} \mid y_0^1),\, \ldots , y_{t-1,t}^{K} \sim \pi_\theta(y_{t-1,t} \mid y_0^K)}
\sum_{k=1}^K \log \\  \frac{
\exp \left( \beta T\, \text{MSE}_t(y^{\tau(k)}_0, y^{\tau(k)}_1; \theta_{\text{ref}}) - \beta T\, \text{MSE}_t(y^{\tau(k)}_0, y^{\tau(k)}_1; \theta_{\text{opt}}) \right)
}{
\sum_{j=k}^K \exp \left( \beta T\, \text{MSE}_t(y^{\tau(j)}_0, y^{\tau(j)}_1; \theta_{\text{ref}}) - \beta T\, \text{MSE}_t(y^{\tau(j)}_0, y^{\tau(j)}_1; \theta_{\text{opt}}) \right)
}
\end{multline}
\endgroup

\subsection{DPO on Flow Matching}\label{sec:dpo}
For completeness, we reproduce the proof from the FlowDPO paper \cite{flowdpo} below.
Given the DPO training objective~\cite{dpo}:
\begin{multline}
    \label{eq:dpo_bt_app}
    \mathcal{L}_{\text{DPO-BT}}(\pi_\theta; \pi_{\text{ref}}) =  - \mathbb{E}_{(x, y_w, y_l) \sim \mathcal{D}} \\
     \log \sigma \big( \beta \log \frac{\pi_\theta(y_w|x)}{\pi_{\text{ref}}(y_w|x)}
     - \beta \log \frac{\pi_\theta(y_l|x)}{\pi_{\text{ref}}(y_l|x)} \big) \Big],
\end{multline}
where $\pi_{\mathrm{opt}},\pi_{\mathrm{ref}}$ are the probabilities produced by the fine-tuned model $\pi_{\mathrm{opt}}$ and the frozen reference model respectively.

It is non-trivial to evaluate $\pi(x)$ for samples derived from flow matching due to the iterative integration.
Following~\cite{wallace2024diffusion}, we discretize the time interval into $T$ steps ($t=i/T$) and rewrite \eqref{eq:dpo_bt_app} as
\begin{multline}
\label{eq:dpo_path}
\mathcal{L}_{\mathrm{DPO}}
  =-\mathbb{E}_{(x, y_w, y_l) \sim \mathcal{D}}\Big[ \\
     \log\sigma \big(
       \beta    
       \mathbb{E}_{x^{w}_{1{:}T},x^{l}_{1{:}T}}
       \!\Bigl[
          \log\frac{\pi_\theta(x^{w}_{0{:}T})}
                    {\pi_{\mathrm{ref}}(x^{w}_{0{:}T})}
        - \log\frac{\pi_\theta(x^{l}_{0{:}T})}
                    {\pi_{\mathrm{ref}}(x^{l}_{0{:}T})}
       \Bigr]
     \big).
\end{multline}

\noindent
Sampling the whole path is expensive, so Jensen’s inequality
bounds~\eqref{eq:dpo_path} by
\begin{multline}
\mathcal{L}_{\mathrm{DPO}}
  \le -\mathbb{E}_{x^{w},x^{l}}
      \log\sigma \\
      \Bigl[
        B\bigl(
           \log\tfrac{\pi_\theta(x_{i-1}^{\,w}\!\mid x_{i}^{w})}
                     {\pi_{\mathrm{ref}}(x_{i-1}^{\,w}\!\mid x_{i}^{w})}
         - \log\tfrac{\pi_\theta(x_{i-1}^{\,l}\!\mid x_{i}^{l})}
                     {\pi_{\mathrm{ref}}(x_{i-1}^{\,l}\!\mid x_{i}^{l})}
        \bigr)
      \Bigr],
\label{eq:dpo_jessen}
\end{multline}
where $B=\beta T$.

Because $(x_{i-1},x_i)$ at an arbitrary step $i$ still cannot be sampled directly, we have to employ the Gaussian paths $p$ in flow matching:
\begin{multline}
\mathcal{L}_{\mathrm{DPO}}
 = -\mathbb{E}_{x^{w},x^{l},i}
    \log\sigma 
    \Bigl(
       B\,\mathbb{E}_{\substack{p(x^{w}_{i-1}\mid x^{w}_i,x^{w}_0)
                               p(x^{l}_{i-1}\mid x^{l}_i,x^{l}_0)}}\\
       \bigl[
         \log\tfrac{p_{\mathrm{opt}}(x_{i-1}^{\,w}\!\mid x_{i}^{w})}
                   {p_{\mathrm{ref}}(x_{i-1}^{\,w}\!\mid x_{i}^{w})} 
       - \log\tfrac{p_{\mathrm{opt}}(x_{i-1}^{\,l}\!\mid x_{i}^{l})}
                   {p_{\mathrm{ref}}(x_{i-1}^{\,l}\!\mid x_{i}^{l})}
       \bigr]
    \Bigr),
    \\
\end{multline}
With Gaussian assumption, we have
\begin{multline}
\mathcal{L}_{\mathrm{DPO}} = -\mathbb{E}_{x^{w},x^{l},i}
    \log\sigma 
    \Bigl(
       B\bigl[
          \mathcal{J}(x_{i}^{w};p,p_{\mathrm{ref}})
        - \mathcal{J}(x_{i}^{w};p,p_{\mathrm{opt}}) \\
        - \mathcal{J}(x_{i}^{l};p,p_{\mathrm{ref}})
        + \mathcal{J}(x_{i}^{l};p,p_{\mathrm{opt}})
       \bigr]
    \Bigr),                                                 \label{eq:flowdpo_gaussian}
\end{multline}
where 
$\mathcal{J}(x_{i};p,p_{\theta})
      \triangleq D_{\mathrm{KL}}\!\bigl(
         p(x_{i-1}\!\mid x_{i}) \,\|\, p_{\theta}(x_{i-1}\!\mid x_{i})
      \bigr)$.
When both $p$ and $p_{\theta}$ are Gaussian with an identical noise
schedule, the divergence reduces to
\begin{equation}
\mathcal{J}(x_{i};p,p_{\theta})
   =\frac{1}{2\sigma_{i-1\mid i}^{2}}
     \bigl\lVert
       \mu(x_{i-1}\!\mid x_{i};0)-\mu_{\theta}(x_{i-1}\!\mid x_{i})
     \bigr\rVert_{2}^{2}.
\label{eq:flowdpo_divergence}
\end{equation}

\noindent
According to DDIM~\cite{ddim}, if the forward path obeys
$x_i\!\sim\!\mathcal{N}(x_i;x_0,\sigma_i^2\mathbf I)$, then
$p(x_{i-1}\!\mid x_i,x_0)=\mathcal{N}\bigl(x_{i-1};
            \mu(x_{i-1}\!\mid x_i,x_0),\sigma_{i-1\mid i}^2\mathbf I\bigr)$
with
\begin{multline}
\mu(x_{i-1}\!\mid x_i,x_0)
   =\frac{1}{\sigma_i}\sqrt{\sigma_{i-1}^2-\sigma_{i-1\mid i}^2}\,x_i \\
    +\Bigl(
        k_{i-1}-\frac{k_i}{\sigma_i}\sqrt{\sigma_{i-1}^2-\sigma_{i-1\mid i}^2}
     \Bigr)x_0.
\end{multline}

Note that all three lines of Gaussian paths, i.e., Variance Preserving (VP), Variance Exploding (VE) and Optimal Transport (OT) meet this requirement.
Replacing $\mathcal{J}$ by an MSE surrogate on the predicted
$x_0$ (or equivalently the noise/velocity) yields the practical loss
\begin{multline}
    \mathcal{L}_{\mathrm{DPO}}
  =-\mathbb{E}_{x^{w}_0,x^{l}_0,x^{w}_i,x^{l}_i}
     \log\sigma \\
     \Bigl(
        B\bigl[
        \mathrm{MSE}_{\xi}(x^{w}_0,x^{w}_i;\theta_{\mathrm{ref}})
        - \mathrm{MSE}_{\xi}(x^{w}_0,x^{w}_i;\theta_{\mathrm{opt}}) \\
        - \mathrm{MSE}_{\xi}(x^{l}_0,x^{l}_i;\theta_{\mathrm{ref}})
        + \mathrm{MSE}_{\xi}(x^{l}_0,x^{l}_i;\theta_{\mathrm{opt}})
        \bigr]
     \Bigr).
\end{multline}

\section{Distributional Metrics}
According to AlphaFlow~\cite{alphaflow}, we introduce the distributional metrics adopted in Table 1 as follows.
Ensemble flexibility is quantified by the mean pairwise C$\alpha$-RMSD between all conformational pairs. 
Atomic-level fluctuations were assessed using both global RMSF (Root Mean Square Fluctuation) and per-residue RMSF; for these, the Pearson correlation coefficient ($r$) with ground-truth values derived from MD simulations was reported to evaluate the accurate reproduction of relative flexibility patterns across the protein structure.
The fidelity of atomic positional distributions was gauged by the root-mean Wasserstein-2 distance (RMWD).
Given two conformational ensembles, $\mathcal{X}$ and $\mathcal{Y}$, and denoting $P(\mathbf{r}_k | \mathcal{E})$ as the empirical 3D coordinate distribution of the $k$-th atom within an ensemble $\mathcal{E}$ comprising $N_{\text{atoms}}$ atoms, the RMWD is defined as:
$\text{RMWD}(\mathcal{X}, \mathcal{Y}) = \sqrt{\frac{1}{N} \sum_{i=1}^N \mathcal{W}_2^2\left(\mathcal{N}[\mathcal{X}_i], \mathcal{N}[\mathcal{Y}_i]\right)}$,
where $\mathcal{N}[\mathcal{X_{i}}]$ are 3D-Gaussians fit to the positional distribution of the ith atom in ensemble X.
Finally, collective motions were evaluated using the MD PCA $\mathcal{W}_2$ distance, calculated as the $\mathcal{W}_2$ distance between conformational distributions projected onto the first two principal components (PCs) obtained from a reference all-atom MD simulation.

Metrics in the Fast-Folding experiments include Jensen-Shannon (JS) distances computed on pairwise $\alpha$-carbon distances (PwD), radius-of-gyration (Rg), and time-lagged independent components (TIC, TIC-2D).
Additionally, we report RMSE$_{\text{contact}}$, which assesses the accuracy of generated ensembles in recovering known protein conformations by aligning sampled conformations to a reference structure and measuring the corresponding root-mean-square deviation (RMSD) of $\alpha$-carbon atoms.
Finally, we quantify structural variability (diversity) within generated ensembles using mean pairwise RMSD, termed RMSF.

\section{Additional Experimental Results}
We provide three case studies for ablation studies of SDE vs. ODE strategies during online sampling in Figure~\ref{fig:ablation_sdeode_appendix}, namely, PNHP, APWF, and SFCH. 

\newpage

%%%%%%%%%%%%%%%%%%%%%%%%%%%%%%%%%%%%%%%%%%%%%%%%%%%%%%%%%%%
\bibliography{aaai2026}
\end{document}